\pdfoutput=1
\documentclass[10pt,twocolumn,letterpaper]{article}
\usepackage{cvpr}
\usepackage{times}
\usepackage{epsfig}
\usepackage{graphicx}
\usepackage{amsmath}
\usepackage{amssymb}
\usepackage{multirow}
\usepackage{amsmath,amssymb}
\usepackage[ruled,vlined]{algorithm2e}

\newcommand{\best}[1]{{\bf \color{red} #1}}
\newcommand{\sbest}[1]{{\bf \color{blue} #1}}
\newcommand{\venue}[1]{{\scriptsize #1}}
\cvprfinalcopy % *** Uncomment this line for the final submission

\def\cvprPaperID{***} % *** Enter the CVPR Paper ID here

\begin{document}

%%%%%%%%% TITLE
\title{Let Features Decide for Themselves: Feature Mask Network for Person Re-identification}

\author{$^1$Guodong Ding,  $^2$Salman Khan, $^1$Zhenmin Tang, $^2$Fatih Porikli\\
$^1$Nanjing University of Science and Technology, $^2$Australian National University\\
{\tt\small gd.ding.cs@gmail.com, salman.khan@anu.edu.au, Tzm.cs@njust.edu.cn, fatih.porikli@anu.edu.au }
}

\maketitle
%\thispagestyle{empty}

%%%%%%%%% ABSTRACT
\begin{abstract}
Person re-identification aims at establishing the identity of a pedestrian from a gallery that contains images of multiple people obtained from a multi-camera system. Many challenges such as occlusions, drastic lighting and pose variations across the camera views, indiscriminate visual appearances, cluttered backgrounds, imperfect detections, motion blur, and noise make this task highly challenging. While most approaches focus on learning features and metrics to derive better representations, we hypothesize that both local and global contextual cues are crucial for an accurate identity matching. To this end, we propose a Feature Mask Network (FMN) that takes advantage of ResNet high-level features to predict a feature map mask and then imposes it on the low-level features to dynamically reweight different object parts for a locally aware feature representation. This serves as an effective attention mechanism by allowing the network to focus on local details selectively. Given the resemblance of person re-identification with classification and retrieval tasks, we frame the network training as a multi-task objective optimization, which further improves the learned feature descriptions. We conduct experiments on Market-1501, DukeMTMC-reID and CUHK03 datasets, where the proposed approach respectively achieves significant improvements of $5.3\%$, $9.1\%$ and $10.7\%$  in mAP measure relative to the state-of-the-art.

\end{abstract}

\begin{figure}
    \centering
    \includegraphics[scale = 0.25]{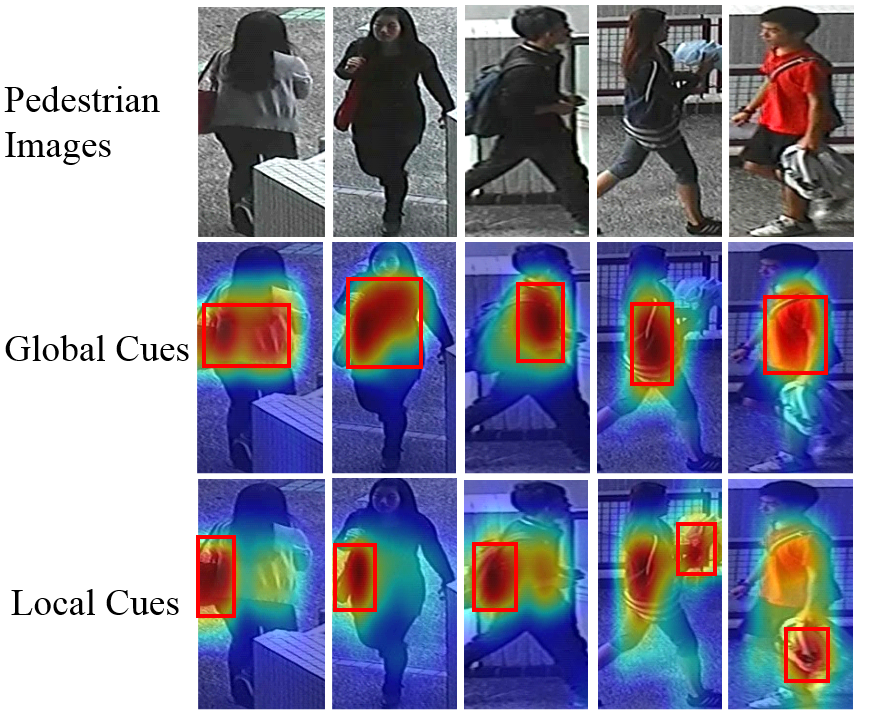}
    \caption{While global information such as the appearance of person's clothes and body shape can be useful for re-identification in some cases, local details such as backpack, handbag, clothing-style and shoes can be even more helpful in other cases. We propose to automatically learn both global and local features specific to the target person identity using the Feature Mask Network.}
    \label{fig:intro}
\end{figure}
%Difficult situations in person re-identification are when pedestrians do look similar. Common identification methods would only provide the global information, while more details such as hairstyle, specific shoes, and a backpack can prove to be more useful. We propose to automatically learn both global and the local features specific to a person's identity using a deep network.%
%While global information such as the background features, appearance of person's clothes and body shape can be useful for re-identification in some cases, more local details such as the hair-style, specific shoes, head-coverings and a backpack can prove to be more useful in other cases. We propose to automatically learn both the global and the local features specific to a person's identity using a deep network.%
%%%%%%%%% BODY TEXT
\section{Introduction}

Person re-identification deals with the task of matching identities across images captured from disjoint camera views. Given a query image, a person re-identification system determines whether the person has been observed by another camera at another time. Significant attention has been dedicated to person re-identification in the past few years, as the task is essential in video surveillance for cross-camera tracking, multi-camera event detection, and pedestrian retrieval. Albeit highly important, the re-identification problem poses significant challenges due to the viewpoint and pose changes, illumination variations, cluttered backgrounds, occlusions, and indiscriminative appearances across different cameras (or even for the same camera). Moreover, re-identification task considers new identities at test time, therefore it requires a high generalization capability of the learned feature encodings.

Existing research on person re-identification problem can be broadly divided into two mainstreams. \textbf{(a)} Methods that seek to learn a discriminative metric, which allows instances of the same identity to be closer while instances of different identities to be far away \cite{craft, guillaumin2009you,koestinger2012large, yi2014deep, wang2016joint,hermans2017defense}.  These metrics learning methods mainly adopt pairwise \cite{guillaumin2009you, koestinger2012large, yi2014deep} or triplet \cite{wang2016joint, hermans2017defense, schroff2015facenet} loss to obtain an embedding for each probe image and distinguish identities in the projected space. Along similar lines, \cite{Chen/cvpr2017} proposes a quadruplet ranking loss that is capable of achieving smaller intra-class variations and large inter-class distances, which results in an improved performance on the test set. \textbf{(b)} Methods that focus on designing robust visual descriptors to model the appearance of the person \cite{matsukawa2016hierarchical, farenzena2010person, zhao2013unsupervised, liao2015person}. Among these techniques, handcrafted features found initial success \cite{farenzena2010person,matsukawa2016hierarchical}. More recently, automatically learned feature representation using deep architectures have shown excellent improvements \cite{zheng2016person,su2016deep,xiao2016learning,zheng2017pedestrian}.
%
%Popular descriptors learning approaches solve this problem with three consecutive steps: 1) train a network same as any classification task with provided id labels on the training dataset, but the main difference between them is that, in person re-id categories in testing set comprise of completely unseen and un-overlapping classes from the training set, 2) exploit top fully-connected layer outputs as final descriptors for both query and gallery images, and 3) calculate similarities based on distance measurements and output the sorted list.
%
The prevalent re-identification approaches belonging to these two research streams assume that the person bounding boxes are provided by a dedicated detector. However, such detections are not always perfect, resulting in problems such as the inclusion of excessive background in the object box, incomplete coverage of body and localization mismatch. This is exacerbated by the fact that there exist heavy occlusions partially masking the pedestrians in surveillance scenarios. 

Our intuition is that a desired capability that allows overcoming these challenges is to pay attention to important yet perhaps subtle local details alongside the supposedly prominent global cues. In this paper, we propose an automatic approach which learns to focus on local details as well as the global image descriptions using deep neural networks. This helps the identification algorithm to filter out the irrelevant image parts and concentrate on the regions that carry more valuable and discriminative cues for the identity prediction task.

We formulate the proposed feature selection strategy as a soft-attention with in a deep network, which enables an end-to-end learning framework. In addition to avoiding the problems due to imperfect pedestrian detection windows, our network learns to resolve ambiguities (such as similar clothing of two different identities) by shifting attention towards more distinguishing aspects of the respective identities. To this end, we utilize already learned global discriminative features as a guidance and a dynamic selection mechanism to assign different importance weights to local feature representations. 

Recent deep learning based re-identification approaches perform training on the identity classification task and use the network features for the test set to perform retrieval \cite{zheng2017pedestrian}. This approach limits the representation capability of deep features since the end-task is different from the one used during learning. To overcome this shortcoming, we propose a multi-task loss formulation that considers both classification and ranking objectives during the training phase. The ranking loss enforces the locally attentive network outputs to take guidance from the predictions based on global features by introducing a margin violation penalty. Our results on three large-scale datasets demonstrate significant performance improvements.   

Our main contributions are threefold, as given below:
\begin{itemize}\setlength{\itemsep}{0em}
\item We propose a Feature Mask Network (FMN) that can dynamically attend to local details in an image and use them alongside global representations for improved pedestrian re-identification. 
\item We introduce a multi-task formulation, which optimizes a classification as well as a pair-wise ranking loss to learn highly robust feature descriptions.
\item The proposed approach is easy to implement, efficient to train, while consistently outperforming the state-of-art methods on all three benchmark datasets.
\end{itemize}
%However, no feature selection or filtering strategy is applied in most existing re-id methods and all parts of feature map share the same importance contribution to the final image representation. In person re-id, when there's excessive backgrounds, it would be helpful if we are able to have some pre-knowledge to help emphasize on parts which would be beneficial to final re-identification, and suppress parts that are not relevant. To this end, we take advantage of already learned discriminative feature as a guidance to give different weights to different parts of feature, functioning as dynamical selection. 

%Compared with existing Reid approaches, our approaches can achieve the state-of-art performance on most large-scale datasets.

The rest of this paper is organized as follows. Section 2 reviews and analyzes the related literature. Section 3 provides the details of our proposed network. Section 4 reports experimental results, and Section 5 concludes this paper with an outlook towards the future work.
%-------------------------------------------------------------------------
\begin{figure*}[!t]
    \centering
    \includegraphics[scale=0.5]{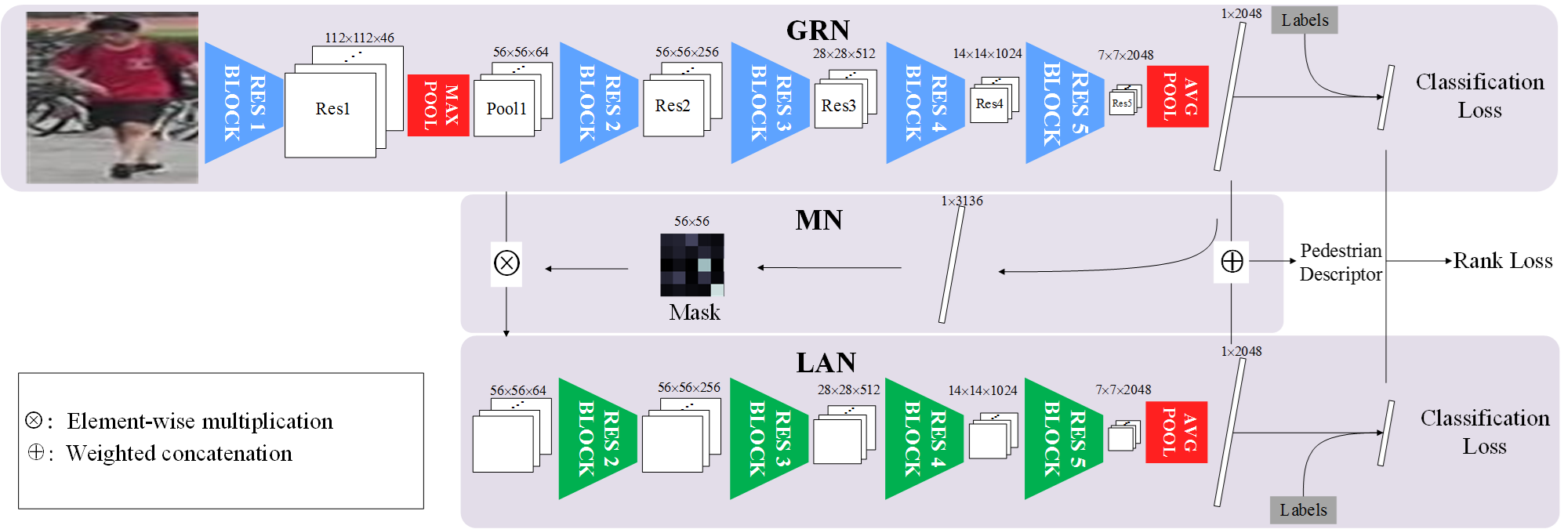}
    \caption{The overall architecture of our network. The GRN obtains high-level feature representations using a fine-tuned ResNet architecture. Using these features, the MN creates a feature mask which dynamically attends to local details useful to person identification. The re-weighted features are used in the bottom LAN to learn more locally focused features. Note that the GRN and LAN learn a separate set of parameters. In testing, we concatenate the last fully connected layer features from both branches to form the final pedestrian descriptor. }
    \label{fig:overall}
\end{figure*}

%-------------------------------------------------------------------
\section{Related Work}
Person re-identification methods employ handcrafted or automatically learned feature representations. There have been several types of handcrafted features used for person re-identification, including local binary patterns (LBP) \cite{koestinger2012large, xiong2014person}, color histogram \cite{koestinger2012large} and local maximal occurrence (LOMO) \cite{liao2015person} statistics. Methods based on handcrafted features focus more on learning an effective distance metric to compare the features, while deep learning based methods jointly learn the best feature representations and the associated distance metrics in a data-driven fashion. We outline the representative works in both streams below.

{\bf Metric learning for person re-identification:}
Several works aim towards metric learning for person re-identification. The main goal is to learn a similarity metric to map similar identities closer in a projected space, while push away different identities as far as possible. This type of work usually takes image pairs or triplets as inputs and learn a similarity metric by adding pairwise or triplet comparison constraints \cite{cheng2016person, varior2016siamese, yi2014deep}. Specifically, \cite{cheng2016person} divides convolution features according to different spatial locations and uses a triplet loss. \cite{yi2014deep} proposed a Siamese network which takes in image pairs and split a pedestrian image into three horizontal parts followed by a separate training process for each part. A cosine distance measure is used  in \cite{yi2014deep} to score similarity of the input images while \cite{chopra2005learning} uses a Euclidean distance. Deep RE-ID \cite{li2014deepreid} integrates several different layers designed for different sub-tasks, such as displacement and pose transforms. \cite{varior2016siamese} utilizes long short-term memory (LSTM) to sequentially process image parts such that the relationships between spatial parts are learned to improve network's discriminative ability.
\cite{wang2016joint} proposes a framework to learn single and cross-image representations using two different CNNs for improving matching performance, and extend their model by utilizing two different CNNs for joint SIR and CIR learning based on either pairwise comparison or triplet objective.

{\bf Deep learning for person re-identification:} Several recent approaches have adopted deep architectures due to their superior expressive power for discriminative feature learning from large-scale visual data \cite{zheng2017pedestrian,wang2016joint}. Some of them focus on learning the image representations together with the similarity function \cite{li2014deepreid, xiao2016learning, wang2016joint}. Other works manage to learn feature representations, tackling person re-identification as a classification problem. Common practices include first training the network to correctly classify samples into different categories and then exploit fully connected layer output as final image descriptor for retrieval \cite{zheng2016person,su2016deep,xiao2016learning,chen2017person}. Zheng \etal \cite{zheng2017unlabeled} adopted generative adversarial networks (GAN) to generate unlabeled data to enlarge the training set. Another group of approaches in the literature incorporate a margin-based loss (e.g., a constrastive loss), which enforces separation between positive and negative pairs \cite{yi2014deep,hermans2017defense}.
Recently, \cite{zheng2016person} showed that a CNN itself can learn very discriminative representations without any extra part-matching or margin-based training. 
Following \cite{zheng2016person}, here we show that a fine tuned network can act as a strong baseline and the proposed FMN can effectively boost its performance.   

{\bf Connection with existing work:} Closest to our proposed approach is the recent pedestrian alignment network by Zheng \etal  \cite{zheng2017pedestrian}. However, their approach is still significantly different to ours since they aim to solve the misalignment problem in automatically detected pedestrian bounding boxes. This misalignment is reduced by applying an affine transformation whose parameters are learned automatically within a network. In contrast, our work advocates that the intermediate feature representations within the network have different levels of relevance for our end-task since the object of interest often only covers a part of the input image. Therefore, directly considering the local features can result in a better performance without the need to align and transform the input image. In our work, we show that appropriately shifting the network's attention towards the local information in feature encoding can greatly help the person re-identification task. Furthermore, the proposed multi-task formulation helps in learning discriminative features by considering the predictions at both global and local levels.  

%While ours focus more on the fact that convention CNNs treat different parts of the feature map equally which is not good enough when target object only covers part of the image input, so we propose to use discriminative features to predict a mask, thus providing different weights for different parts of feature maps to further improve re-identification performance.

\section{Proposed Method}

Our approach is based on the proposition that a successful person re-identification system needs to give importance to both the global and the local discriminative aspects of a pedestrian whose image is acquired from different views using multiple surveillance cameras. To this end, we introduce a novel CNN based deep learning architecture, which learns to focus both on the global and local cues of a person that are useful for its re-identification. We describe our proposed architecture and training procedure in the following sections. 

\subsection{Feature Mask Network}
The proposed CNN architecture is shown in Fig.~\ref{fig:overall}. The complete system consist of three main components, which we term as (from \emph{top} to \emph{bottom})  \textbf{a)} Global Representation Network, \textbf{b)} Mixing Network and \textbf{c)} Locally Attentive Network. 

The Global Representation Network (GRN) learns the holistic feature representations corresponding to an input image. It is designed as a Residual Network \cite{he2016deep} with five residual blocks each (except the first module) containing skip connections. The GRN has a total of 50 parameter layers with a total of 3.8 billion FLOPs. It has been pre-trained on the ImageNet image classification dataset and fine-tuned on the person re-identification dataset. The Mixing Network (MN) predicts the mask weights for the local features from the initial layers in the GRN. These weights are derived from the global feature representation learned using the GRN for the pedestrian images. The MN consists of a transformation layer implemented as a fully connected (FC) layer on top of the global features from GRN, followed by a reshaping module and a mixer which performs element-wise product between the local feature representations and the mask weights. 

The input pedestrian images may contain excessive background or mis-alignment errors due to high appearance, scale and pose variations in the candidate profiles. Therefore, the important information in an image may get suppressed in the global representation learned by the GRN. The third block in our scheme, called the Locally Attentive Network (LAN), learns to attend to local discriminative features which can provide useful clues for a person's identity matching. A LAN takes local features from the GRN, which are re-weighted by the mask predicted using the MN. These modified activations basically exhibit an attention mechanism, where the discriminating local features are given more importance compared to others. Furthermore, the proposed attention mechanism also effectively incorporates the relevant contextual information which is useful for the person re-identification task. The LAN consists of four residual blocks with similar architecture to the corresponding blocks in the GRN. Since, the LAN parameters are not shared with the GRN, it learns a different global representation with a refocus on the locally discriminating information. This information acts as a complementary source of information which we found to be highly useful in our experiments (see Sec.~\ref{subsec:Evaluation}). In the following, we describe the details of mask computation.

\begin{algorithm}[h]
\KwIn{Pre-trained model $\phi(\cdot,\boldsymbol{\theta_o})$,
    Re-ID training data $I$, 
    Identity labels $Y$,
    Maximum Iterations $T_g$, $T_l$}
\KwOut{Learnt FMN Model $\phi(\cdot,\boldsymbol{\hat{\theta_g}}; \boldsymbol{\hat{\theta_m}} , \boldsymbol{\hat{\theta_l}})$,  $\boldsymbol{\hat{\theta_g}}$, $\boldsymbol{\hat{\theta_m}}$ and $\boldsymbol{\hat{\theta_l}}$ denote parameters of GRN, MN and LAN, respectively.} 
{\bf Initialization:} $\boldsymbol{\theta_o} \rightarrow \boldsymbol{\theta_g}$, $\boldsymbol{\theta_o}\rightarrow \boldsymbol{\theta_l}$, random initialization for $\boldsymbol{\theta_m}$ \\ 
{\bf Global representation learning}\;
\nl \For{$\boldsymbol{t = 1:T_g}$}{
	Keep $\boldsymbol{\theta_m}$, $\boldsymbol{\theta_l}$ fixed\;
	Update $\boldsymbol{\theta_g^t}$ using Eq. \eqref{cls}}
\nl $\hat{\boldsymbol{\theta_g}} \leftarrow \boldsymbol{\theta_g^{T_g}}$ \;
{\bf Locally attentive representation learning}\;
\nl \For{$\boldsymbol{t = 1:T_l}$}{
	Keep $\boldsymbol{\hat{\theta_g}}$ fixed and feed-forward GRN\;
    Produce mask $\boldsymbol{m^t}$ using Eq. \eqref{fc} and Eq. \eqref{mask}\;
    Apply mask on LAN inputs with $\boldsymbol{m^t}$ using Eq. \eqref{apply}\; 
    Update $\boldsymbol{\theta_m^t}$ and $\boldsymbol{\theta_l^t}$ using both Eq. \eqref{cls} and Eq. \eqref{rank}\;
	}
\nl $\hat{\boldsymbol{\theta_m}} \leftarrow \boldsymbol{\theta_m^{T_l}}$, $\hat{\boldsymbol{\theta_l}} \leftarrow \boldsymbol{\theta_l^{T_l}}$ \;
{\bf Return:} $\phi(\cdot,\boldsymbol{\hat{\theta_g}},\boldsymbol{\hat{\theta_m}},\boldsymbol{\hat{\theta_l}})$\textbf{}
    
    \caption{{\bf Feature Mask Net Optimization} \label{Algorithm}}

\end{algorithm}

\subsection{Mask Computation}
The MN operates on the global feature representation $\mathbf{g} \in \mathbb{R}^{m}$ from higher layers (final fully connected layer in our case)  and the local feature  representation $\mathbf{f} \in \mathbb{R}^{n}$ from lower layers (output from first residual block in our case) in the Residual network. Since the feature representations from the lower layers in a CNN is arranged as multiple 2D activation maps for color images, we can represent their dimensions more conveniently as: $n = h\times w\times c$, where $h$, $w$ and $c$ denote the height, width and the number of feature channels respectively. The MN first transforms the global feature from GRN to a $n' = h\times w$ dimensional output which can then be used to compute the feature mask:
\begin{align}
\label{fc}
 \mathbf{m}' = \sigma(\mathbf{W}^{\text{T}}\mathbf{g}),
\end{align}
where, $\sigma$ denotes the ReLU activation function and $\mathbf{m}'\in \mathbb{R}^{n'}$ is the transitional mask and $\mathbf{W}\in \mathbb{R}^{m\times n'}$ is the weight matrix for the transformation which is equivalent to a fully connected layer in our mixing network. As a result, we can learn this feature mapping directly from the data which can provide an image-specific mask for the local features. Since our goal is to attend to local features in the spatial domain, we identically re-weight all feature channels in $\mathbf{f}$ using the same predicted feature mask.  The final feature mask ($\mathbf{m} \in \mathbb{R}^{h\times w}$) is obtained from $\mathbf{m}'$ by reshaping and applying element-wise exponentiation as follows: 
\begin{align}
\label{mask}
\mathbf{m}_{i,j} &= \exp(\mathbf{m}_{k}), \notag\\
 s.t.,\;  j &= \lfloor k/h \rfloor + 1,\, i = k - h(j-1)
\end{align}
Once the feature mask $\mathbf{m}$ is obtained, MN uses a mixer to combine it with $\mathbf{f}$ using the channel-wise Hadamard product (denoted as `$\circ$' below):
\begin{align}
\label{apply}
\mathbf{o}_{i} = \mathbf{f}_{i}\circ\mathbf{m}, \quad s.t.,\, i \in [1,c]
\end{align}
where, $i$ denotes the feature channel number in $\mathbf{f}$. We outline the training process for our proposed network below.

\subsection{Classification and Ranking}
The network is trained in two stages, summarized in Algorithm~\ref{Algorithm}. \emph{First}, the GRN is trained to predict the pedestrian identities  in an end-to-end manner. The parameter learning process for GRN involves a weight initialization step using a ResNet model pretrained on the ImageNet dataset, followed by our task-specific fine-tuning. Once the training process is complete, the feature representation from the GRN encodes global discriminator information corresponding to a given image. \emph{Afterwards}, the GRN weights are kept fixed, while the MN and LAN weights are learned jointly in the next stage. Similar to the first stage, the second stage training is also performed using the pedestrian identities as the ground-truth labels. In contrast to the global representations learned via GRN, the second stage training focuses on the locally discriminate information and re-shifts the attention appropriately using the MN to obtain a complementary feature representation. 

Both the GRN and LAN are trained for the identity classification task using supervised learning in two stages. We use a conventional cross-entropy loss function for both stages as follows:
\begin{equation}
\label{cls}
\mathcal{L}_{cls}(\mathbf{p}, \mathbf{y}) = - \sum\limits_k y_k \log \frac{\exp(p_k)}{\sum_j \exp(p_j) }, \qquad k \in [1,r] 
\end{equation}
where $\mathbf{p} \in \mathbb{R}^{r}$ denote the predicted output and $\mathbf{y} \in \mathbb{R}^{r}$ denotes the desired output as a one-hot vector. Here, $r$ denotes the number of classes in the dataset, which is equal to the total units in the output layer.

However, the softmax loss does not directly consider the ranking errors. Therefore, in the second stage of joint network training, we add on top of LAN a ranking loss defined as follows :
\begin{equation}
\label{rank}
\mathcal{L}_{rank}(p^{G}_{t},p^{L}_t) = \max(0, p^{G}_t-p^{L}_t+m),
\end{equation}
where $m$ represents the margin, $p^{G}_t,p^{L}_t$ denote the prediction probability  on the correct category label $t$ of GRN and LAN, respectively. Imposing rank loss enables LAN to take the prediction form GRN as reference, and enforces LAN to make better predictions for correct labels by a margin, thus leading to more confident and accurate predictions.

\subsection{Image Descriptor} 
After learning the parameters of the network, the final image descriptor at test time is obtained by combining the feature representations from the GRN and the LAN.  These feature representations are derived from the last FC layer in each network, which contain task specific discriminative information pertaining to both global and local pedestrian attributes. 
The following relation is used to compute a re-weighted concatenation of normalized individual descriptors:

\begin{equation}
D = [\alpha \frac{\mathbf{g}^\text{T}}{\parallel \mathbf{g}\parallel_2}, (1-\alpha) \frac{\mathbf{l}^\text{T}}{\parallel \mathbf{l}\parallel_2} ]^\text{T},
\end{equation}
where $\parallel\cdot\parallel_2$ operator denotes an $\ell^2$-norm. The parameter $\alpha$ decides the trade-off between features $\mathbf{g}, \mathbf{l}$ from the two separate breaches, GRN and LAN, respectively. We set $\alpha$ to $0.5$ in our experiments following \cite{zheng2017pedestrian}. The resulting descriptor is used to find closest matches from the gallery by performing a nearest neighbour (NN) search based on Euclidean distance. This provides an initial ranking list, which is further improved by the re-ranking process, as described below.

\subsection{Re-ranking}\label{sec:re-rank}
The person re-identification process can be viewed as a retrieval task, in which a given query image is used to retrieve the images containing the same identity. Based on the initial ranking obtained using the descriptor $D$, we perform a re-ranking step to further improve the re-identification performance. The re-ranking step discovers the relationships in the initial ranking to remove the false matches and obtain an improved list. For example, a simple strategy is to remove the matches which do not conform with the top-$k$ ranked images \cite{ye2016person}. However, this approach performs poorly when the top-$k$ matches are noisy. To overcome this problem, we use the $k$-reciprocal nearest neighbours in the re-ranking step as proposed in \cite{zhong2017re}. By definition, two images are reciprocal nearest neighbours if a search using one of the images ranks the second one among the top-$k$ images. This reduces the false positives in the re-ranked list and results in significantly better performance. 

The re-ranking process operates in an unsupervised manner.  Specifically, a $k$-reciprocal feature ($D_r$) is calculated for each image using the gallery by computing reciprocal neighbouring relationships \cite{zhong2017re}. Given a query image, this feature is used along side the descriptor $D$  to find similarity with the gallery. Note that instead of the Euclidean distance, the Jaccard similarity measure is used to match $k$-reciprocal features. The final distance is calculated by the aggregation of both distance measures. Remarkably, the re-ranking process relies heavily on the features calculated using our proposed network architecture. A flawed feature representation can lead to a degraded performance as a result of re-ranking. In our case, a boost is performance using the re-ranking approach shows the strength of our proposed feature description (see Sec.~\ref{subsec:Evaluation}).

\section{Experiments}
We extensively evaluate the proposed approach on three person re-identification datasets: Market-1501  \cite{zheng2015scalable}, CUHK03 \cite{li2014deepreid}, and DukeMTMC-reID \cite{ristani2016performance}. We first briefly provide the dataset details and the evaluation protocol, followed by our experimental results and analysis.

% \usepackage{multirow}

\begin{table*}
\begin{center}
%\scriptsize
{\tabcolsep=0.88mm
\begin{tabular}{|l|c|cccc|cccc|cccc|cccc|}
\hline
\multirow{2}{*}{Methods} & \multicolumn{1}{l|}{\multirow{2}{*}{dim}} & \multicolumn{4}{c|}{Market-1501} & \multicolumn{4}{c|}{DukeMTMC-reID} & \multicolumn{4}{c|}{CUHK03 detected} & \multicolumn{4}{c|}{CUHK03 labeled} \\ \cline{3-18} 
 & \multicolumn{1}{l|}{} & 1 & 5 & 20 & mAP & 1 & 5 & 20 & mAP & 1 & 5 & 20 & mAP & 1 & 5 & 20 & mAP \\ \hline\hline
GRN & 2048 & 79.33 & 91.48 & 96.62 & 58.50 & 67.91 & 81.33 & 89.59 & 48.40 & 38.00 & 48.14 & 59.93 & 33.68 & 34.36 & 46.64 & 58.14 & 30.14 \\
LAN & 2048 & 81.15 & 91.39 & 96.44 & 59.88 & 69.84 & 83.03 & 90.04 & 51.05 & 31.50 & 46.07 & 60.86 & 29.39 & 32.36 & 47.57 & 61.93 & 30.41 \\ \hline
FMN & 4096 & {\bf 85.99} & {\bf 93.74} & {\bf 97.51} & {\bf 67.12} & {\bf 74.51} & {\bf 85.05} & {\bf 92.41} & {\bf 56.88} & {\bf 42.57} & {\bf 56.21} & {\bf 67.36} & {\bf 39.21} & {\bf 40.71} & {\bf 54.57} & {\bf 65.50} & {\bf 38.05} \\ \hline 
\end{tabular}}
\end{center}
\caption{Comparison of different methods on Market-1501, DukeMTMC-reID, CUHK03 (detected), and CUHK03 (labeled). Rank-1, 5, 20 accuracy (\%) and mAP (\%) are reported. Consistent improvement of our proposed on all datasets in terms of rank@k accuracy and mAP can be observed.}
\label{tab:overall}
\end{table*}

\subsection{Datasets}
\begin{figure}[!t]
    \centering
    \includegraphics[scale=0.2]{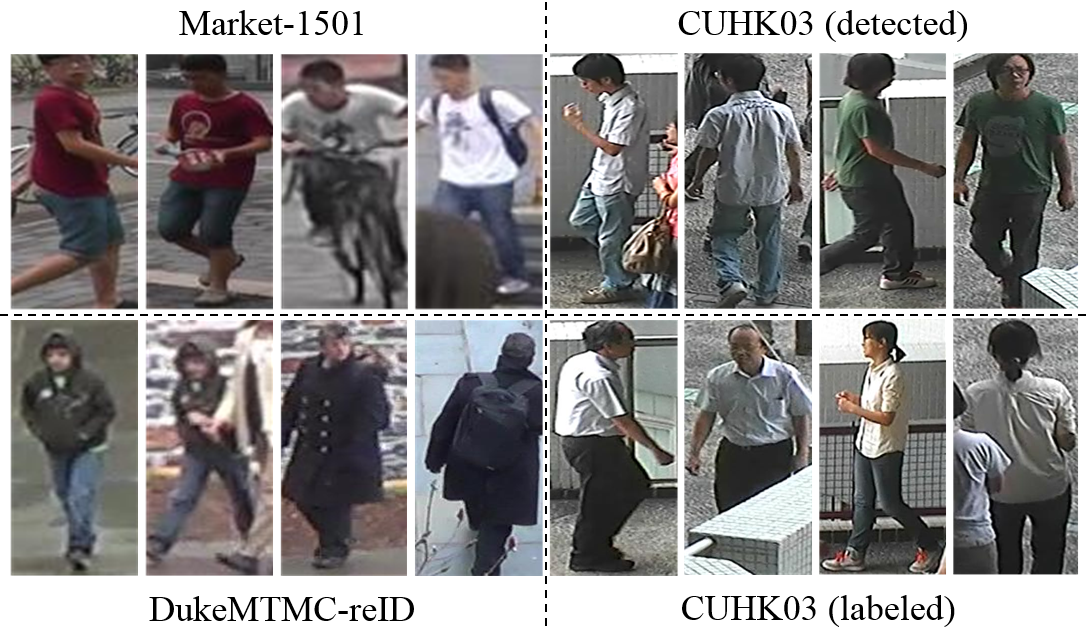}
    \caption{Pedestrian image samples from different datasets. In each grid, we show two identities with two captured images each. As we can see, there are illumination variances, view point changes and even large occlusions.%, which make this task more challenging.  }
    }
    \label{fig:dataset}
\end{figure}
{\bf Market-1501:} Market-1501 is a person re-identification dataset consisting of 32,688 images with bounding boxes of 1501 pedestrians. These image are captured on campus and persons are detected using Deformable Parts Model (DPM) \cite{felzenszwalb2008discriminatively}. Each identity has at most been captured by 6 different cameras. In this experiment setting, 12,936 detected images of 751 identities been taken as training set, whilst the rest 19,732 images of 750 identities are used for testing, following the original evaluation protocols.

{\bf DukeMTMC-reID:} DukeMTMC-reID is one of the largest pedestrian image datasets derived from DukeMTMC \cite{ristani2016performance} and comprises of 36,411 pedestrian images of 1,404 identities. It is split into train/test sets of 16,522/19,889, and we evaluate with 2,228 queries to retrieve from 17,661 gallery images. The dataset introduces occlusion of pedestrians, such as cars or trees, which makes this task more challenging.

{\bf CUHK03:} CUHK03 contains 14,097 images of 1,467 identities. On average, each identity has 9.6 images captured by 5 different sets of cameras. The dataset provides both manually cropped bounding boxes and the automatically cropped bounding boxes using a pedestrian detector, named as `\emph{CUHK03 labeled}' and `\emph{CUHK03 detected}', respectively. For fair comparison, we evaluate our method on the new training/testing protocol proposed in \cite{zhong2017re} which divides original dataset by half, yielding a training set of 767 identities and a testing set of 700 identities. \cite{zhong2017re} argue that this split conforms with the real-world person re-id tasks, which can only provide limited training samples, while re-id is performed on a larger unseen sample pool.

\subsection{Implementation Details}
{\bf Network Training:} Due to the high classification accuracy of ResNet-50 \cite{he2016deep}, following the baseline in \cite{zheng2016person} we also use it as our backbone architecture. The network is pre-trained on the ImageNet dataset consisting of 1000 object classes. To fine-tune it for the person re-identification task, we replace the 1000 units in the final FC layer with the number of training identities in dataset. This forms the GRN branch in our network, which is trained with an initial learning rate of 0.1, that is reduced to 0.01 after 20 epochs. For the MN and LAN branches, we train with the same initial learning rate, which is reduced to 0.01 after 35 epochs. The training is performed until the the network converges. We update our parameters with stochastic gradient descent with 0.9 momentum. The training dataset is augmented with horizontal flipping and cropping the original images.

{\bf Evaluation Metrics:}
We evaluate our methods with mean average precision (mAP) and rank-1, rank-5 and rank-20 accuracy measures. The rank-i accuracy denotes the rate at which one or more correctly matched images appear in top-i. If no correctly matched images appear in the top-i of the sorted list, rank-i=0, otherwise rank-i=1. We report the mean rank-i accuracy for query images. Also, for each method, we calculate the area under the Precision-Recall curve and the mean of the average precision scores for each query, which reflects the overall precision and recall rate.

The proposed framework is implemented using the MatConvNet \cite{vedaldi15matconvnet} library.

\subsection{Evaluation}\label{subsec:Evaluation}

\begin{table}
\begin{center}
\begin{tabular} {|l|c|c|}
\hline 
Method & rank-1 & mAP\\
\hline\hline
DADM \venue{(ECCV'16)} \cite{su2016deep} & 39.4 & 19.6\\
BoW+kissme \venue{(ICCV'15)} \cite{zheng2015scalable} & 44.42 & 20.76\\
MR-CNN \venue{(AVSS'17)} \cite{Ustinova2017MultiregionBC} & 45.58 & 26.11\\
MST-CNN \venue{(ACMMM'16)} \cite{liu2016multi} & 45.10 & -\\
FisherNet \venue{(PR'17)} \cite{wu2017deep} & 48.15 & 29.94\\
CAN \venue{(TIP'17)} \cite{liu2017end} & 48.24 & 24.43\\
SL \venue{(CVPR'16)} \cite{chen2016similarity} & 51.90 & 26.35\\
DNS \venue{(CVPR'16)} \cite{zhang2016learning} & 55.43 & 29.87\\
Gate Reid \venue{(ECCV'17)} \cite{varior2016gated} & 65.88 & 39.55\\
SOMAnet \venue{(ArXiv'17)} \cite{barbosa2017looking} & 73.87 & 47.89\\
Verif,-Identif. \venue{(TOMM'17)} \cite{zheng2016discriminatively} & 79.51 & 59.87\\
MSCAN \venue{(CVPR'17)}\cite{Li_2017_CVPR} & 80.31 & 57.53\\
SVDNet \venue{(ICCV'17)}\cite{Sun_2017_ICCV} & 82.3 & 62.1\\
SSM \venue{(CVPR'17)} \cite{Bai_2017_CVPR} & 83.7 & 65.5\\
GAN \venue{(ICCV'17)} \cite{zheng2017unlabeled} & 83.97 & 66.07\\
APR \venue{(ArXiv'17)} \cite{lin2017improving} & 84.29 & 64.67\\
%Triplet\cite{hermans2017defense} & 84.92 & 69.14\\
%Triplet+re-rank\cite{hermans2017defense} &  86.67 & \best{81.07}\\
PAN \venue{(ArXiv'17)} \cite{zheng2017pedestrian} & 82.81 & 63.35\\
PAN+re-rank \venue{(ArXiv'17)} \cite{zheng2017pedestrian} & 85.78 & \sbest{76.56}\\
JLML \venue{(IJCAI'17)} \cite{li2017person} & 85.1 & 65.5 \\
DPFL \venue{(ICCV'17)} \cite{chen2017person} & \best{88.9} & 73.1 \\
\hline
Basel. & 79.33 & 58.50\\
FMN & 85.99 & 67.12\\
FMN+re-rank & \sbest{87.92} & \best{80.62}\\
\hline
\end{tabular}
\end{center}
\caption{Rank-1 accuracy (\%) and mAP (\%) on Market-1501 dataset. The best and second best performance are marked in \best{red} and \sbest{blue}, respectively.}
\label{tab:market}
\end{table}

{\bf Effectiveness of FMN.} We comprehensively evaluate our proposed FMN on all three  large-scale re-identification benchmarks to show the effectiveness of our proposed network architecture. The individual performance of the GRN, LAN and overall results of the complete model are shown in Table \ref{tab:overall}. This serves as an ablation study. Since GRN is a fine-tuned version of the pre-trained model, therefore GRN is used as a strong baseline in this work. As one can note, independent LAN feature achieves score that's close to the baseline branch, and when combined with the baseline feature, a significant boost in overall performance is observed. Rank-1 accuracy for Market-101, DukeMTMC-reID, CUHK03 detected and CUHK03 labeled datasets have been improved by a margin of 6.66\%, 6.60\%, 4.57\% and 6.35\%, respectively. The performance in terms of mAP values have also been improved remarkably by 8.62\%, 8.48\%,5.53\% and 7.91\%, respectively. This consistent performance gain proves that the locally attentive feature representations learned by the MN and LAN are complementary to the global features from GRN. Notably, the rank-1 LAN accuracy for CUHK03 detected dataset experience a considerable decline (around 6.5\%) compared with the baseline, we speculate that this behaviour is due to the severe occlusions and detection errors in the dataset.

%\begin{figure}
%\centering
%    \includegraphics[scale = 0.38]{images/grnlan}
%    \caption{Visualization of focuses on different parts of GRN and LAN. }
%    \label{fig:grnlan}
%\end{figure}

{\bf Comparison with the State-of-the-Art Methods:} We also compare our proposed approach with the state-of-art methods on Market-1501, DukeMTMC-reID and CUHK03 datasets in Tables \ref{tab:market}, \ref{tab:duke} and \ref{tab:cuhk03} respectively. On all three datasets, we achieve the best or the second best performances in comparison to the very recent methods with more sophisticated pipelines. As shown in Table~\ref{tab:market}, we achieved {\bf rank-1 = 87.92\%, mAP=80.62\%} using the re-ranking approach explained in Sec.~\ref{sec:re-rank}, which is a highly consistent performance across both metrics. On DukeMTMC-reID, our proposed method also achieved the best performance with {\bf rank-1 = 79.52\%, mAP = 72.79\%}. On CUHK03 dataset, we observe our rank-1 is {\bf 5.6\%} higher than best competing methods on detected and {\bf 2.1\%} on labeled datasets. 

%\Nate{Since the figure's been replaced, this might need to be commented out}As shown in Figure \ref{fig:finalres}, some retrieval results on three datasets are demonstrated. Comparing to baseline, the ranks of matches has been greatly improved, true matches receive higher ranks, while false matches get lower ranks.

\begin{figure*}
%    \raggedleft
	\centering
    \includegraphics[scale=0.35]{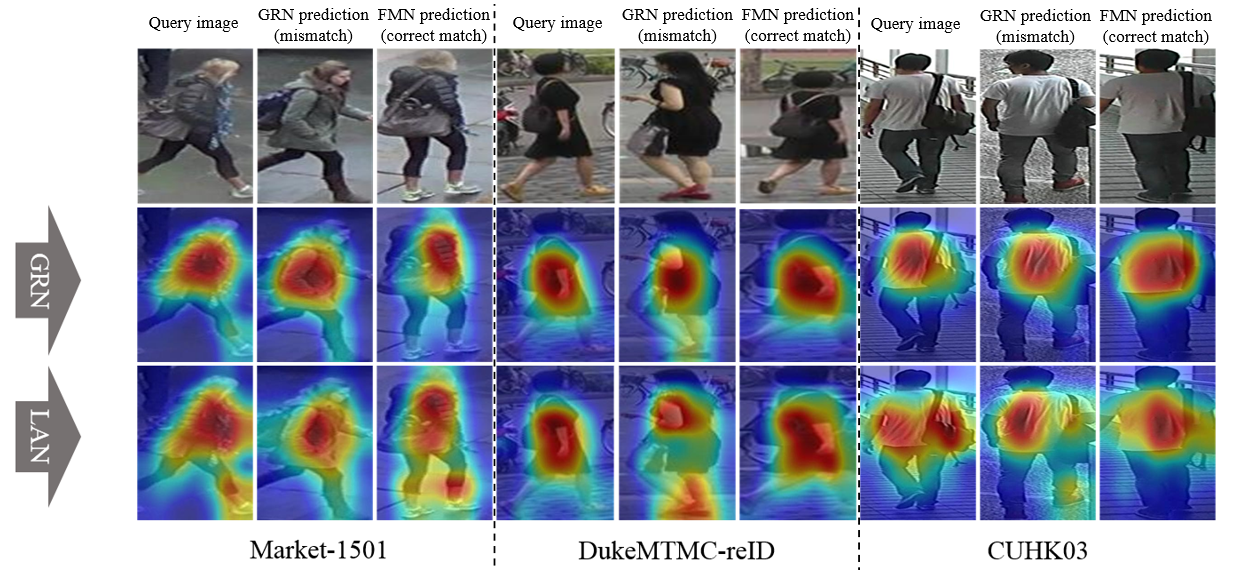}
    \caption{Qualitative examples of improved retrieval results on the Market-150, DukeMTMC-reID and CUHK03 datasets (\emph{left} to \emph{right}). For each query image, we present its false rank-1 match based solely on GRN (denoted by mismatch) followed by its accurate match using our proposed architecture. Second and third rows visualize the heat maps obtained from GRN and LAN, respectively. One can observe that by attending to local distinctive parts (e.g., subtle variations in apparel design, shoes and carry on bags) using LAN, the overall performance is boosted and the proposed network is therefore better suited for re-identification task.}
    \label{fig:finalres}
\end{figure*}

{\bf GRN vs LAN:} Alongside the quantitative comparisons on the three aforementioned datasets, we also qualitatively analyze the effect of FMN for the case if highly confusing pedestrian examples. To this end, we illustrate query images from each of the three datasets in Fig.~\ref{fig:finalres} along with the Rank-1 mismatch (predicted by baseline) and a true match (predicted by our proposed model) for the respective pedestrian. We also visualize the heat maps obtained from both the GRN and LAN to study the salient image regions which are given more attention by the network during the prediction process. One can notice from both Figure\ref{fig:intro} and Figure \ref{fig:finalres} that global representations from GRN focus mainly around the main torso of the body. This can lead to incorrect predictions because the upper trunk of the body can be identical for two altogether different identities. In contrast, more subtle local details such as the clothing and attire specifics (\emph{left} example), footwear (\emph{middle} example) and differences of the back-packs (\emph{right} example) can provide more useful cues for correct identification of persons. The FMN attends to both global and local details and leverages their complementary characteristics to correctly identify the corresponding match for a query image.

\begin{table}
\begin{center}
\begin{tabular}{|l|c|c|}
\hline
Method & rank-1 & mAP\\
\hline\hline
BoW+kissme \venue{(ICCV'15)} \cite{zheng2015scalable} & 25.13 & 12.17\\
LOMO+XQDA \venue{(CVPR'15)} \cite{liao2015person}  & 30.75 & 17.04\\
GAN \venue{(ICCV'17)} \cite{zheng2017unlabeled} & 67.68 & 47.13\\
OIM \venue{(CVPR'17)} \cite{xiao2017joint} & 68.1 & -\\
APR \venue{(ArXiv'17)} \cite{lin2017improving} & 70.7 & 51.9\\
SVDNet \venue{(ICCV'17)} \cite{Sun_2017_ICCV} & 76.7 & 56.8\\
PAN \venue{(ArXiv'17)} \cite{zheng2017pedestrian} & 71.6 & 51.5\\
PAN+re-rank \venue{(ArXiv'17)} \cite{zheng2017pedestrian} & 75.9 & \sbest{66.7}\\
DPFL \venue{(ICCV'17)} \cite{chen2017person} & \sbest{79.2} & 60.6\\
\hline
Basel. & 67.91 & 48.40\\
FMN & 74.51 & 56.88\\
FMN+re-rank & \best{79.52} & \best{72.79}\\
\hline
\end{tabular}
\end{center}
\caption{Rank-1 accuracy (\%) and mAP (\%) on DukeMTMC-reID.}
\label{tab:duke}
\end{table}

\begin{table}[]
\begin{center}\setlength\tabcolsep{2.5pt}
\begin{tabular}{|l|c|c|c|c|}
\hline
\multirow{2}{*}{Method} & \multicolumn{2}{c|}{Detected} & \multicolumn{2}{c|}{Labeled} \\ \cline{2-5} 
 & rank-1 & mAP & rank-1 & mAP \\ \hline\hline
BoW+XQDA \venue{(ICCV'15)} \cite{zheng2015scalable} & 6.4 & 6.4 & 8.0 & 7.3 \\
LOMO+XQDA \venue{(CVPR'15)} \cite{liao2015person}  & 12.8 & 11.5 & 14.8 & 13.6 \\
ResNet+XQDA \venue{(CVPR'17)} \cite{Zhong_2017_CVPR} & 31.1 & 28.2 & 32.0 & 29.6 \\
\cite{Zhong_2017_CVPR}+re-rank \venue{(CVPR'17)} & 34.7 & 37.4 & 38.1 & 40.3 \\
PAN \venue{(ArXiv'17)} \cite{zheng2017pedestrian} & 36.3 & 34.0 & 36.9 & 35.0 \\
PAN+re-rank \venue{(ArXiv'17)} \cite{zheng2017pedestrian} & \sbest{41.9} & \sbest{43.8} & \sbest{43.9} & \sbest{45.8} \\
DPFL \venue{(ICCV'17)} \cite{chen2017person} & 40.7 & 37.0 & 43.5 & 40.5\\\hline
Basel. & 38.2 & 34 & 34.4 & 30.1 \\
FMN & 42.6 & 39.2 & 41.0 & 38.1 \\
FMN+re-rank & \best{47.5} & \best{48.5} & \best{46.0} & \best{47.6} \\ \hline
\end{tabular}
\end{center}
\caption{Results on CUHK03 dataset with new evaluation protocol. New protocol divide each dataset roughly by half for training and testing. Under this setting, we use a larger testing gallery and smaller training set.  }
\label{tab:cuhk03}
\end{table}

{\bf Local Feature Selection:} The GRN extracts a hierarchy of features corresponding to different level of details in the input image. From initial to final layers, the output features transition from local to global information. We evaluate the effect of local feature selection from GRN on the person re-identification accuracy. In Table~\ref{tab:layers}, we report rank-1 accuracy and mAP when the local features from different GRN layers are used in the mixing network. For the case of Res1, we apply masking operation on the output of pooling layer (Pool1) which significantly reduces the MN parameters to only a quarter of that on Res1. We can see a consistent performance boost as the selected layer is changed from final layers towards initial GRN layers. Specifically, the rank-1 accuracy of 82.39 and mAP of 61.95 at Res4 changes to 85.99 and 67.12, respectively, at Pool1. This observation is intuitive, since the network achieves best performance when the complementary information from both global and local levels are combined for the re-identification task. The high layers encode more holistic information about appearance and shape, while the local information can disambiguate cases with high appearance similarities at a global level. When the local features from higher layers are used, the resulting performance is low since the features from near-by layers encode redundant information. 
%\Nate{Should we say something like high layers are more robust yet encode more semantic features while lower convolutional layers carry more discriminative information can help separate cases with similar appearances.}

\begin{table}[!htp]
\begin{center}
\begin{tabular}{|l|c|c|c|}
\hline
 & mask size & rank-1 & mAP \\ \hline\hline
Pool1 & 56$\times$56 & \textbf{85.99} & \textbf{67.12} \\ \hline
Res2 & 56$\times$56 & 85.15 & 65.84 \\ \hline
Res3 & 28$\times$28 & 84.23 & 65.16 \\ \hline
Res4 & 14$\times$14 & 82.39 & 61.95 \\ \hline
\end{tabular}
\end{center}
\caption{Rank-1 accuracy and mAP results with different layer being masked on Market-1501 dataset.}
\label{tab:layers}
\end{table}

%{\bf What would feature select for themselves.} 
%we first investigate the mask itself.
%\todo{Need to add a figure.}To better understand the work behind this feature mask, we visualize the predicted feature mask in fig. First, we extract all output for weight candidates, and then take the exponential to each candidate to obtain non negative numbers. After this step, we squash them to the range of [0,1] to treat them as a scaling factor to each feature map activations.  We observe that with predicted mask, XXX regions has been magnified while xxx regions has been suppressed, which %
%-------------------------------------------------------------------------
\section{Conclusion}
%One can shoot in the heart when he knows where to aim.
A person occupies only a portion of the input image, and a global scene description does not suffice for an accurate identity matching. In this work, we proposed a hybrid architecture for CNN, which simultaneously learns to focus on the more discriminative parts of the input scene. Given a global feature, we directly predict the attention mask which is used to re-weight the local scene details in the feature space. This strategy allows the flexibility to re-focus attention on the local details which can be highly valuable for predicting a persons unique identity. The locally aware feature description leads to a highly compact and complementary feature representation, which together with the global representation achieves highly accurate results on three large-scale datasets. Significant boosts are observed when the proposed features are used along-with a re-ranking strategy, demonstrating the strength of proposed features to correctly encode reciprocal relationships between person identities.  
%In future we will apply the proposed feature mask prediction on Siamese style pairwise training.%

{\small
\bibliographystyle{ieee}
\bibliography{egbib}

\begin{thebibliography}{10}\itemsep=-1pt

\bibitem{Bai_2017_CVPR}
S.~Bai, X.~Bai, and Q.~Tian.
\newblock Scalable person re-identification on supervised smoothed manifold.
\newblock In {\em The IEEE Conference on Computer Vision and Pattern
  Recognition (CVPR)}, July 2017.

\bibitem{barbosa2017looking}
I.~B. Barbosa, M.~Cristani, B.~Caputo, A.~Rognhaugen, and T.~Theoharis.
\newblock Looking beyond appearances: Synthetic training data for deep cnns in
  re-identification.
\newblock {\em arXiv preprint arXiv:1701.03153}, 2017.

\bibitem{chen2016similarity}
D.~Chen, Z.~Yuan, B.~Chen, and N.~Zheng.
\newblock Similarity learning with spatial constraints for person
  re-identification.
\newblock In {\em Proceedings of the IEEE Conference on Computer Vision and
  Pattern Recognition}, pages 1268--1277, 2016.

\bibitem{Chen/cvpr2017}
W.~Chen, X.~Chen, J.~Zhang, and K.~Huang.
\newblock Beyond triplet loss: a deep quadruplet network for person
  re-identification.
\newblock In {\em The Conference on Computer Vision and Pattern Recognition},
  2017.

\bibitem{chen2017person}
Y.~Chen, X.~Zhu, and S.~Gong.
\newblock Person re-identification by deep learning multi-scale
  representations.
\newblock In {\em Proceedings of the IEEE International Conference on Computer
  Vision}, 2017.

\bibitem{craft}
Y.-C. Chen, X.~Zhu, W.-S. Zheng, and J.-H. Lai.
\newblock Person re-identification by camera correlation aware feature
  aug-mentation.
\newblock {\em IEEE Transactions on Pattern Analysis and Machine Intelligence
  (DOI: 10.1109/TPAMI.2017.2666805)}.

\bibitem{cheng2016person}
D.~Cheng, Y.~Gong, S.~Zhou, J.~Wang, and N.~Zheng.
\newblock Person re-identification by multi-channel parts-based cnn with
  improved triplet loss function.
\newblock In {\em Proceedings of the IEEE Conference on Computer Vision and
  Pattern Recognition}, pages 1335--1344, 2016.

\bibitem{chopra2005learning}
S.~Chopra, R.~Hadsell, and Y.~LeCun.
\newblock Learning a similarity metric discriminatively, with application to
  face verification.
\newblock In {\em Computer Vision and Pattern Recognition, 2005. CVPR 2005.
  IEEE Computer Society Conference on}, volume~1, pages 539--546. IEEE, 2005.

\bibitem{farenzena2010person}
M.~Farenzena, L.~Bazzani, A.~Perina, V.~Murino, and M.~Cristani.
\newblock Person re-identification by symmetry-driven accumulation of local
  features.
\newblock In {\em Computer Vision and Pattern Recognition (CVPR), 2010 IEEE
  Conference on}, pages 2360--2367. IEEE, 2010.

\bibitem{felzenszwalb2008discriminatively}
P.~Felzenszwalb, D.~McAllester, and D.~Ramanan.
\newblock A discriminatively trained, multiscale, deformable part model.
\newblock In {\em Computer Vision and Pattern Recognition, 2008. CVPR 2008.
  IEEE Conference on}, pages 1--8. IEEE, 2008.

\bibitem{guillaumin2009you}
M.~Guillaumin, J.~Verbeek, and C.~Schmid.
\newblock Is that you? metric learning approaches for face identification.
\newblock In {\em Computer Vision, 2009 IEEE 12th international conference on},
  pages 498--505. IEEE, 2009.

\bibitem{he2016deep}
K.~He, X.~Zhang, S.~Ren, and J.~Sun.
\newblock Deep residual learning for image recognition.
\newblock In {\em Proceedings of the IEEE conference on computer vision and
  pattern recognition}, pages 770--778, 2016.

\bibitem{hermans2017defense}
A.~Hermans, L.~Beyer, and B.~Leibe.
\newblock In defense of the triplet loss for person re-identification.
\newblock {\em arXiv preprint arXiv:1703.07737}, 2017.

\bibitem{koestinger2012large}
M.~Koestinger, M.~Hirzer, P.~Wohlhart, P.~M. Roth, and H.~Bischof.
\newblock Large scale metric learning from equivalence constraints.
\newblock In {\em Computer Vision and Pattern Recognition (CVPR), 2012 IEEE
  Conference on}, pages 2288--2295. IEEE, 2012.

\bibitem{Li_2017_CVPR}
D.~Li, X.~Chen, Z.~Zhang, and K.~Huang.
\newblock Learning deep context-aware features over body and latent parts for
  person re-identification.
\newblock In {\em The IEEE Conference on Computer Vision and Pattern
  Recognition (CVPR)}, July 2017.

\bibitem{li2014deepreid}
W.~Li, R.~Zhao, T.~Xiao, and X.~Wang.
\newblock Deepreid: Deep filter pairing neural network for person
  re-identification.
\newblock In {\em Proceedings of the IEEE Conference on Computer Vision and
  Pattern Recognition}, pages 152--159, 2014.

\bibitem{li2017person}
W.~Li, X.~Zhu, and S.~Gong.
\newblock Person re-identification by deep joint learning of multi-loss
  classification.
\newblock In {\em {IJCAI}}, pages 2194--2200. ijcai.org, 2017.

\bibitem{liao2015person}
S.~Liao, Y.~Hu, X.~Zhu, and S.~Z. Li.
\newblock Person re-identification by local maximal occurrence representation
  and metric learning.
\newblock In {\em Proceedings of the IEEE Conference on Computer Vision and
  Pattern Recognition}, pages 2197--2206, 2015.

\bibitem{lin2017improving}
Y.~Lin, L.~Zheng, Z.~Zheng, Y.~Wu, and Y.~Yang.
\newblock Improving person re-identification by attribute and identity
  learning.
\newblock {\em arXiv preprint arXiv:1703.07220}, 2017.

\bibitem{liu2017end}
H.~Liu, J.~Feng, M.~Qi, J.~Jiang, and S.~Yan.
\newblock End-to-end comparative attention networks for person
  re-identification.
\newblock {\em IEEE Transactions on Image Processing}, 2017.

\bibitem{liu2016multi}
J.~Liu, Z.-J. Zha, Q.~Tian, D.~Liu, T.~Yao, Q.~Ling, and T.~Mei.
\newblock Multi-scale triplet cnn for person re-identification.
\newblock In {\em Proceedings of the 2016 ACM on Multimedia Conference}, pages
  192--196. ACM, 2016.

\bibitem{matsukawa2016hierarchical}
T.~Matsukawa, T.~Okabe, E.~Suzuki, and Y.~Sato.
\newblock Hierarchical gaussian descriptor for person re-identification.
\newblock In {\em Proceedings of the IEEE Conference on Computer Vision and
  Pattern Recognition}, pages 1363--1372, 2016.

\bibitem{ristani2016performance}
E.~Ristani, F.~Solera, R.~Zou, R.~Cucchiara, and C.~Tomasi.
\newblock Performance measures and a data set for multi-target, multi-camera
  tracking.
\newblock In {\em European Conference on Computer Vision}, pages 17--35.
  Springer, 2016.

\bibitem{schroff2015facenet}
F.~Schroff, D.~Kalenichenko, and J.~Philbin.
\newblock Facenet: A unified embedding for face recognition and clustering.
\newblock In {\em Proceedings of the IEEE Conference on Computer Vision and
  Pattern Recognition}, pages 815--823, 2015.

\bibitem{su2016deep}
C.~Su, S.~Zhang, J.~Xing, W.~Gao, and Q.~Tian.
\newblock Deep attributes driven multi-camera person re-identification.
\newblock In {\em European Conference on Computer Vision}, pages 475--491.
  Springer, 2016.

\bibitem{Sun_2017_ICCV}
Y.~Sun, L.~Zheng, W.~Deng, and S.~Wang.
\newblock Svdnet for pedestrian retrieval.
\newblock In {\em The IEEE International Conference on Computer Vision (ICCV)},
  Oct 2017.

\bibitem{Ustinova2017MultiregionBC}
E.~Ustinova, Y.~Ganin, and V.~S. Lempitsky.
\newblock Multiregion bilinear convolutional neural networks for person
  re-identification.
\newblock In {\em AVSS}, 2017.

\bibitem{varior2016gated}
R.~R. Varior, M.~Haloi, and G.~Wang.
\newblock Gated siamese convolutional neural network architecture for human
  re-identification.
\newblock In {\em European Conference on Computer Vision}, pages 791--808.
  Springer, 2016.

\bibitem{varior2016siamese}
R.~R. Varior, B.~Shuai, J.~Lu, D.~Xu, and G.~Wang.
\newblock A siamese long short-term memory architecture for human
  re-identification.
\newblock In {\em European Conference on Computer Vision}, pages 135--153.
  Springer, 2016.

\bibitem{vedaldi15matconvnet}
A.~Vedaldi and K.~Lenc.
\newblock Matconvnet -- convolutional neural networks for matlab.
\newblock In {\em Proceeding of the {ACM} Int. Conf. on Multimedia}, 2015.

\bibitem{wang2016joint}
F.~Wang, W.~Zuo, L.~Lin, D.~Zhang, and L.~Zhang.
\newblock Joint learning of single-image and cross-image representations for
  person re-identification.
\newblock In {\em Proceedings of the IEEE Conference on Computer Vision and
  Pattern Recognition}, pages 1288--1296, 2016.

\bibitem{wu2017deep}
L.~Wu, C.~Shen, and A.~van~den Hengel.
\newblock Deep linear discriminant analysis on fisher networks: A hybrid
  architecture for person re-identification.
\newblock {\em Pattern Recognition}, 65:238--250, 2017.

\bibitem{xiao2016learning}
T.~Xiao, H.~Li, W.~Ouyang, and X.~Wang.
\newblock Learning deep feature representations with domain guided dropout for
  person re-identification.
\newblock In {\em Proceedings of the IEEE Conference on Computer Vision and
  Pattern Recognition}, pages 1249--1258, 2016.

\bibitem{xiao2017joint}
T.~Xiao, S.~Li, B.~Wang, L.~Lin, and X.~Wang.
\newblock Joint detection and identification feature learning for person
  search.
\newblock In {\em Proc. CVPR}, 2017.

\bibitem{xiong2014person}
F.~Xiong, M.~Gou, O.~Camps, and M.~Sznaier.
\newblock Person re-identification using kernel-based metric learning methods.
\newblock In {\em European conference on computer vision}, pages 1--16.
  Springer, 2014.

\bibitem{ye2016person}
M.~Ye, C.~Liang, Y.~Yu, Z.~Wang, Q.~Leng, C.~Xiao, J.~Chen, and R.~Hu.
\newblock Person reidentification via ranking aggregation of similarity pulling
  and dissimilarity pushing.
\newblock {\em IEEE Transactions on Multimedia}, 18(12):2553--2566, 2016.

\bibitem{yi2014deep}
D.~Yi, Z.~Lei, S.~Liao, and S.~Z. Li.
\newblock Deep metric learning for person re-identification.
\newblock In {\em Pattern Recognition (ICPR), 2014 22nd International
  Conference on}, pages 34--39. IEEE, 2014.

\bibitem{zhang2016learning}
L.~Zhang, T.~Xiang, and S.~Gong.
\newblock Learning a discriminative null space for person re-identification.
\newblock In {\em Proceedings of the IEEE Conference on Computer Vision and
  Pattern Recognition}, pages 1239--1248, 2016.

\bibitem{zhao2013unsupervised}
R.~Zhao, W.~Ouyang, and X.~Wang.
\newblock Unsupervised salience learning for person re-identification.
\newblock In {\em Proceedings of the IEEE Conference on Computer Vision and
  Pattern Recognition}, pages 3586--3593, 2013.

\bibitem{zheng2015scalable}
L.~Zheng, L.~Shen, L.~Tian, S.~Wang, J.~Wang, and Q.~Tian.
\newblock Scalable person re-identification: A benchmark.
\newblock In {\em Proceedings of the IEEE International Conference on Computer
  Vision}, pages 1116--1124, 2015.

\bibitem{zheng2016person}
L.~Zheng, Y.~Yang, and A.~G. Hauptmann.
\newblock Person re-identification: Past, present and future.
\newblock {\em arXiv preprint arXiv:1610.02984}, 2016.

\bibitem{zheng2016discriminatively}
Z.~Zheng, L.~Zheng, and Y.~Yang.
\newblock A discriminatively learned cnn embedding for person
  re-identification.
\newblock {\em TOMM}, 2017.

\bibitem{zheng2017pedestrian}
Z.~Zheng, L.~Zheng, and Y.~Yang.
\newblock Pedestrian alignment network for large-scale person
  re-identification.
\newblock {\em arXiv preprint arXiv:1707.00408}, 2017.

\bibitem{zheng2017unlabeled}
Z.~Zheng, L.~Zheng, and Y.~Yang.
\newblock Unlabeled samples generated by gan improve the person
  re-identification baseline in vitro.
\newblock In {\em Proceedings of the IEEE International Conference on Computer
  Vision}, 2017.

\bibitem{zhong2017re}
Z.~Zhong, L.~Zheng, D.~Cao, and S.~Li.
\newblock Re-ranking person re-identification with k-reciprocal encoding.
\newblock 2017.

\bibitem{Zhong_2017_CVPR}
Z.~Zhong, L.~Zheng, D.~Cao, and S.~Li.
\newblock Re-ranking person re-identification with k-reciprocal encoding.
\newblock In {\em The IEEE Conference on Computer Vision and Pattern
  Recognition (CVPR)}, July 2017.

\end{thebibliography}
}

\end{document}